\documentclass[12pt]{iopart}
\expandafter\let\csname equation*\endcsname\relax
\expandafter\let\csname endequation*\endcsname\relax
\usepackage[cmex10]{amsmath}
\usepackage[font=footnotesize]{subfig}
\usepackage{amssymb}
\usepackage[dvips]{graphicx}
\usepackage{fixltx2e}
\usepackage[abbr]{harvard}
\usepackage[export]{adjustbox}
\usepackage{textcomp}

\DeclareMathOperator{\DSC}{DSC}

\begin{document}
\let\WriteBookmarks\relax
\def\floatpagepagefraction{1}
\def\textpagefraction{.001}

\title{Deep morphological recognition of kidney stones using intra-operative endoscopic digital videos}
\author{Vincent Estrade$^{1}$, Michel Daudon$^{2}$, Emmanuel Richard$^{3}$, Jean-Christophe Bernhard$^{1}$, Franck Bladou$^{1}$, Gregoire Robert$^{1}$, Laurent Facq$^{4}$, Baudouin Denis de Senneville$^{4}$}
\address{$^1$ Department of Urology, CHU Pellegrin, Place Amélie Raba Léon 33000 Bordeaux, France}
\address{$^2$ Department of Multidisciplinary Functional Explorations, AP-HP, Tenon Hospital, INSERM UMRS 1155, Sorbonne University, Paris, France}
\address{$^3$ University of Bordeaux, INSERM, BMGIC, U1035, CHU Bordeaux, 33076 Bordeaux, France}
\address{$^4$ University of Bordeaux, CNRS, INRIA, Bordeaux INP, IMB, UMR 5251, F-33400 Talence, France}



 
 



\begin{abstract}
The collection and the analysis of kidney stone morphological criteria are essential for an aetiological diagnosis of stone disease. However, \textit{in-situ} LASER-based fragmentation of urinary stones, which is now the most established chirurgical intervention, may destroy the morphology of the targeted stone. In the current study, we assess the performance and added value of processing complete digital endoscopic video sequences for the automatic recognition of stone morphological features during a standard-of-care intra-operative session.

To this end, a computer-aided video classifier was developed to predict \textit{in-situ} the morphology of stone using an intra-operative digital endoscopic video acquired in a clinical setting. The proposed processing pipeline included the following four steps: 1) a neural network identified relevant stone regions on each frame; 2) a quality control (QC) module ensured the presence of (potentially fragmented) stones and sufficient image stability throughout recording; 3) a second neural network predicted the stone morphologies; 4) a final module performed the endoscopic stone recognition (ESR) from all collected morphological observations. 

The proposed technique was evaluated on pure (\emph{i.e.} include one morphology) and mixed (\emph{i.e.} include at least two morphologies) stones involving ``Ia/Calcium Oxalate Monohydrate (COM)'', ``IIb/ Calcium Oxalate Dihydrate (COD)'' and ``IIIb/Uric Acid (UA)'' morphologies. 585 static images were collected (349 and 236 observations of stone surface and section, respectively) and employed for artificial intelligence (AI)-training. 71 digital endoscopic videos (50 exhibited only one morphological type and 21 displayed two) were analyzed using the proposed video classifier (56840 frames processed in total). 
Using the proposed approach, diagnostic performances (averaged over both pure and mixed stone types) were as follows: balanced accuracy=$[88 \pm 6]$ (min=81) $\% $, sensitivity=$[80 \pm 13]$ (min=69) $\%$, specificity=$[95 \pm 2]$ (min=92) $\%$, precision=$[78 \pm 12]$ (min=62) $\%$ and F1-score=$[78 \pm 7]$ (min=69) $\%$.

The obtained results demonstrate that AI applied on digital endoscopic video sequences is a promising tool for collecting morphological information during the time-course of the stone fragmentation process without resorting to any human intervention for stone delineation or selection of good quality steady frames. To this end, irrelevant image information must be removed from the prediction process at both frame and pixel levels, which is now feasible thanks to the use of AI-dedicated networks. 

\end{abstract}

\vspace{2pc}

\maketitle

\section{Introduction}

Retrograde Intrarenal Surgery (RIRS) with flexible ureteroscopes and LASER is now the most established chirurgical intervention for urinary stones \cite{1}. This is made possible thanks to the recent technological development of: 

\begin{itemize}

 \item Endoscopic devices allowing intra-operative ``live'' imaging sessions \cite{2} \cite{3}.
 
  \item LASER devices allowing \textit{in-situ} stone fragmentation such as Holmium-Yag, which operates in ``popcorn'' \cite{4} or ``dusting'' modes \cite{5}, or more recently Thulium Fiber LASER (TFL) \cite{6} \cite{7}).
 
 \item  Surgical materials designed to the convenient collection and extraction of residual stone fragments (clamp or small basket \cite{8}).

\end{itemize}

During this interventional process, the collection and the analysis of stone morphological criteria are essential for an aetiological diagnosis of stone disease \cite{2} \cite{9}. Seven groups (denoted by Roman numerals I, II, ..., VII) are currently distinguished by the international morpho-constitutional classification of urinary stones. Each group, which is associated with a specific crystalline type, is then divided into subgroups (denoted by a subscript in the Latin alphabet attached to the Roman numeral: Ia, Ib, ...) that differentiate morphologies and aetiologies for a given crystalline type. Morphological stone observations can be collected post-operatively by examining extracted stone fragments using binocular magnifying glass \cite{10} \cite{11} and spectrophotometric infrared recognition (FTIR) \cite{10} \cite{11} \cite{12}. However, it has been reported that LASER fragmentation of stones may destroy the morphology of the targeted stone \cite{13} \cite{14}. Fortunately, it has been recently shown that morphological observations may be collected \textit{in-situ} during an intra-operative endoscopic inspection \cite{9}. Importantly, while the most recent (chronologically) lithogenic events are located on the surface of the stone, less recent events are observable on a section, and initial lithogenic context are observable in the nucleus of the stone \cite{9} \cite{13}. Morphological aspects of entire stones may thus be collected both before and during the time-course of the stone fragmentation process using Endoscopic Stone Recognition (ESR) \cite{2} \cite{3} \cite{9}. 

\begin{figure}[h!]
\begin{minipage}[b]{\linewidth}
\centering
\centerline{\includegraphics[width=\linewidth]{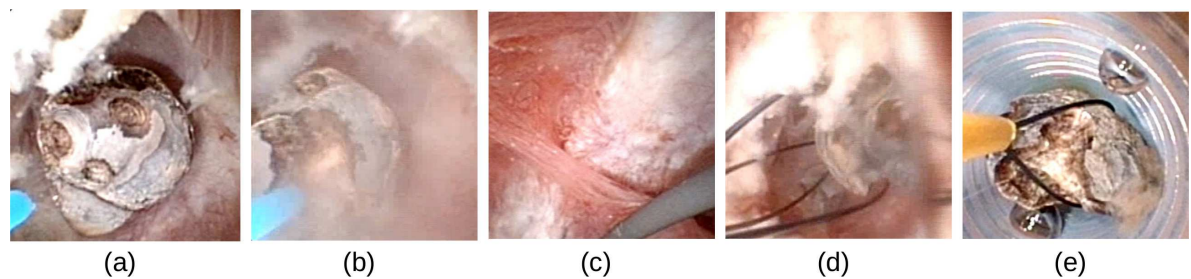}}
\end{minipage}
\caption{Typical events recorded by the endoscope during the time-course of a LASER fragmentation session. (a): stone surface examination (visual characteristics of the Ia/COM morphology are observable smooth or mammillary/dark-brown surface), (b): stone fragmentation,  (c): stone free prospection, (d): use of a clamp, (e): removal of a stone fragment. A basket is used in (d) and (e) to extract stone fragments.}
\label{fig:Endoscopic_Session}
\end{figure}

It has been reported that a relatively fast learning curve is needed to acquire the ESR skill for the most frequently encountered stone morphologies: calcium oxalate monohydrate (COM, also referred to as Ia by the above-mentioned international morpho-constitutional classification of urinary stones), calcium oxalate dihydrate (COD or IIa/IIb) and uric acid (UA or IIIa/IIIb) \cite{15}. However, a much steeper learning curve is mandatory when mixed stones morphologies are involved (\emph{i.e.} include at least two morphologies, which cover almost half of cases \cite{9} \cite{15}). This issue may hamper the clinical adoption of ESR \cite{2} \cite{9}.

Numerous studies demonstrated that ESR may be conveniently supported by artificial intelligence (AI) algorithm, which appears as a promising asset for automatic ESR (AESR) of the morphological components using intra-operative digital images \cite{3} \cite{9} \cite{16} \cite{17} \cite{18} \cite{19}, even when both pure and mixed stones are involved \cite{3} \cite{9}. However AI-based ESR currently relies on the manual selection of a good quality steady frame from video by an endo-urologist trained in identifying stone morphology. An inherent human-selection bias arises: it is likely that the images chosen by the trained endo-urologist would contain the necessary visual cues needed for morphologic identification. Thus, it is not obvious that static images collected by someone unfamiliar with stone morphological types would be as useful. Moreover, AI networks, which are trained on a selection of annotated endoscopic images, hardly generalize when images are acquired dynamically on-the-fly during the common practice of an intra-operative endoscopic imaging session. Such images are typically disturbed by motion, particles flying around in the saline or dusting event, as well as various specular reflections or scene illumination variations. AI-based ESR can be further complicated by the presence in the image field-of-view of the tip of the endoscope or surgical materials needed for the extraction of residual stone fragments (clamp/small basket) or by stone-free prospection sessions. Several typical events recorded by the endoscope during the time-course of a LASER fragmentation session are reported in figure \ref{fig:Endoscopic_Session}.

\begin{figure}[h!]
\begin{minipage}[b]{\linewidth}
\centering
\centerline{\includegraphics[width=\linewidth]{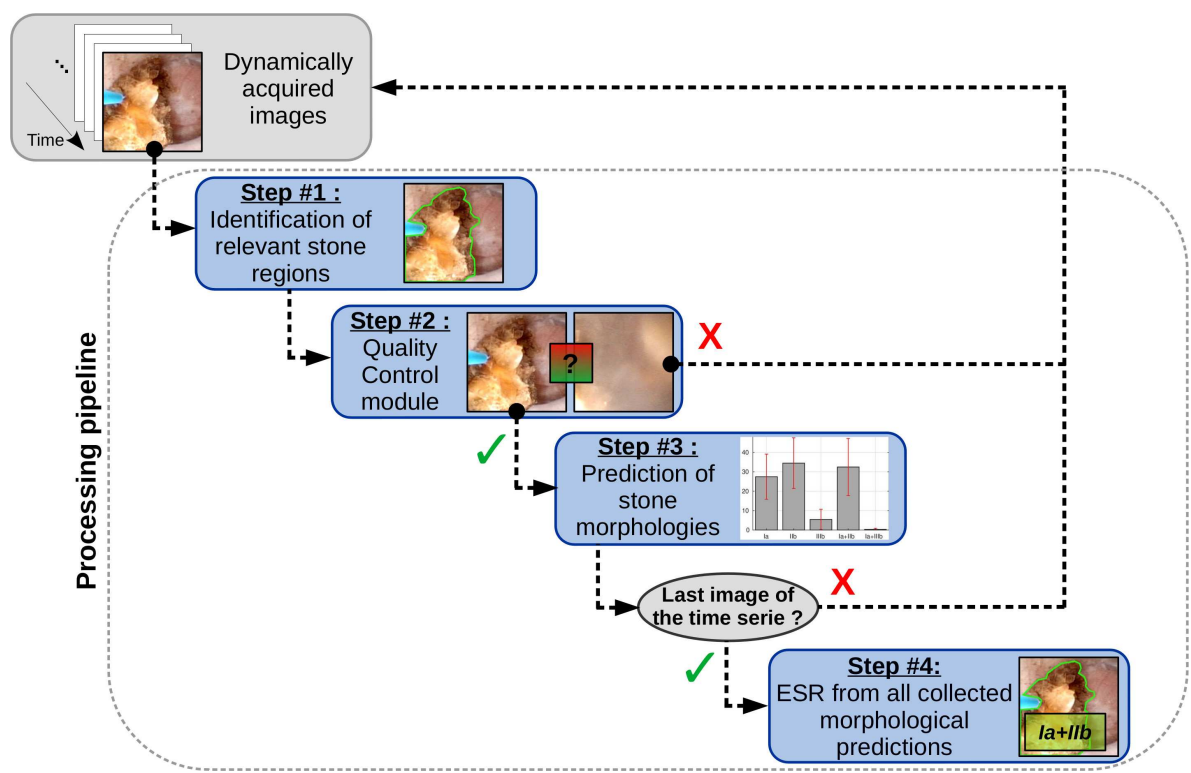}}
\end{minipage}
\caption{Processing pipeline proposed to perform an automatic ESR using an intra-operative digital endoscopic video as input. The four successive processing steps are detailed in the blue blocs.}
\label{fig:Pipeline}
\end{figure}

The contribution of the current study is two-fold:

\begin{enumerate}

\item A fully automatic classifier is proposed to perform AESR based on intra-operative endoscopic video sessions without resorting to any human intervention for stone delineation or selection of good quality steady frames. The processing pipeline, which is summarized in figure \ref{fig:Pipeline}, includes the following four successive steps: 1) a dedicated neural network segmented relevant stone regions on each frame; 2) a quality control (QC) module ensures both the presence of a stone and a sufficient image stability; 3) a second neural network predicts stone morphologies; 4) a module analyzes morphological observations collected during the complete duration of the video to produce the final predicted stone type.
 
\item The added value of digital intra-operative endoscopic videos for intra-operative AESR is assessed on the most frequently encountered stone morphologies (Ia/COM, IIb/COD and IIIb/UA). Besides, we evaluate the benefit of collecting morphological information revealed during the time-course of the stone fragmentation process without resorting to any human intervention for stone delineation or selection of good quality steady frames from videos.

\end{enumerate}

To the best of our knowledge, this study assesses for the first time the performance and added value of processing complete digital endoscopic video sequences for the automatic recognition of stone morphological features during a standard-of-care intra-operative session.

\section{Materials and Methods}

\subsection{Study design}

A urologist (VE, 20 years of experience) intra-operatively and prospectively collected endoscopic digital images and videos of stones encountered between January 2018 and November 2020 in a single centre (CHU Pellegrin, Bordeaux). A flexible digital ureterorenoscope (Olympus URF-V CCD sensor) was employed. The study adhered to all local regulations and data protection agency recommendations (National Commission on Data Privacy requirements). Patients were informed that their data would be used anonymously.

For each collected image and video, morphological criteria were collected and classified by the trained urologist (VE) according to recommendations outlined in \cite{9}. Endoscopic stone recognition was confirmed by observations of LASER-fragmented stones obtained using microscopy (binocular magnifying glass) and spectrophotometric infrared recognition (Fourier Transform InfraRed spectroscopy), as proposed in \cite{9}. 
Only data involving the following five morphological classes were considered in this study:

\begin{itemize}

 \item Three pure stone morphologies: we limited to most common morphologies that urologist encounter, namely Ia/COM, IIb/COD and IIIb/UA morphologies.
 
 \item Two mixed stone morphologies: we limited to the following mixed stones composed of two morphologies: Ia/COM + IIb/COD and Ia/COM + IIIb/UA.

\end{itemize}

\subsection{High quality image database for AI-training}

High-quality endoscopic images were collected during intra-operative sessions to build up training cohorts designed to feed the neural networks involved in this study. Endoscopic images were collected with the stone intact (so called ``surface'' images), and after the stone is broken to reveal its interior (so called ``section'' images). The stability of the endoscopic video image for certain duration, the absence of stone dust and is a prerequisite to ensure an adequate image quality. Any motion event is likely to hamper the image quality and, in turn, bias the outcome of the trained network. The trained urologist (VE) had to make several attempts to get sharp images (2 attempts in average, max = 4). All images were cropped and resampled to a common dimension of $256 \times 256$ pixels.

\subsection{Video database for AESR validation}

Intra-operative videos were recorded (separate patients than those used for the above-mentioned collection of high quality images). Collected videos were used to compute morphological stone prediction scores of the proposed computer-aided ESR method. To reduce the amount of data and for data standardization considerations, the videos were temporally resampled to a common frame-rate of 8 Hz. All frames were subsequently cropped and resampled to a common dimension of $256 \times 256$ pixels.

\subsection{Computer-aided AESR using intra-operative digital endoscopic videos}

Several image regions, which are not necessary present during AI-training (surrounding tissue, tip of the endoscope, surgical material as well as various unpredictable floating fragments), need to be discarded in the computer-aided AESR process. To this end, each frame of in the video database was analysed (in chronological order of video recording) using the processing pipeline summarized in figure \ref{fig:Pipeline}, which includes the following four steps:

\paragraph{Step $\#$1: Frame-wise segmentation of relevant stone regions}

\begin{figure}[h!]
\begin{minipage}[b]{\linewidth}
\centering
\centerline{\includegraphics[width=\linewidth]{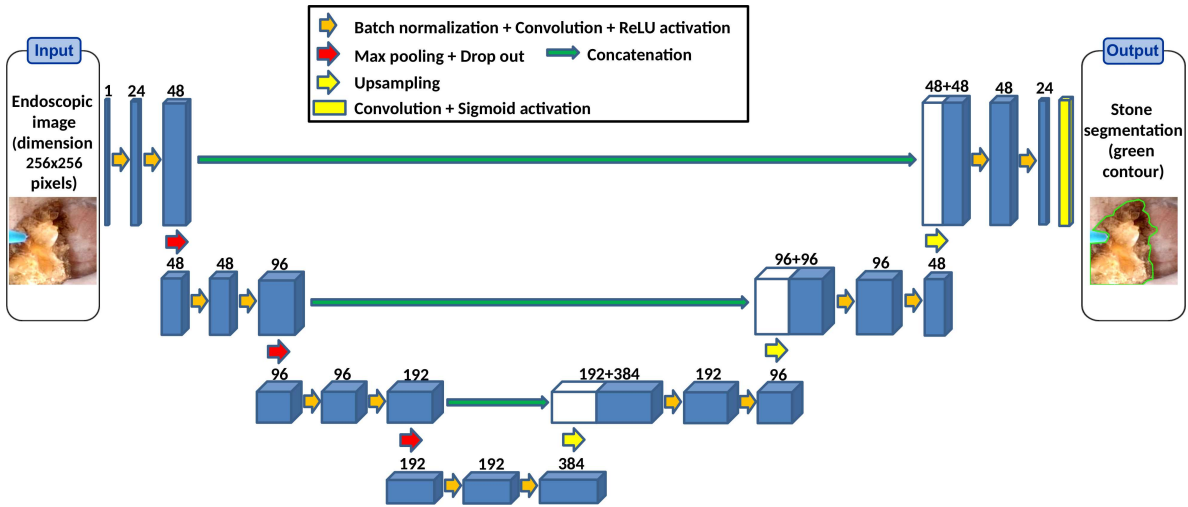}}
\end{minipage}
\caption{Illustration of the used U-Net architecture for the segmentation of the kidney stone. A dynamically acquired endoscopic image (dimension $256 \times 256$ pixels) is used as a single input channel. Each block of the CNN (blue rectangle) is composed of batch normalization, convolution and ReLU activation. The number of $3 \times 3$ filters is indicated on the top of each block.} 
\label{fig:SegCNN}
\end{figure}

A dedicated neural network identified relevant stone regions on each frame. The training cohort used to feed the neural network was composed by representative images manually selected in the above-mentioned high-quality intra-operative image database. On each image, a medical physicist blinded to the participants’ characteristics manually delineated kidney stones. These manual segmentations were considered as the ground truth. A 2D-convolutional neural network (CNN) was trained utilizing this training cohort. The 2D CNN architecture used for the segmentation of a kidney stone in a single frame of the endoscopic video is presented in figure \ref{fig:SegCNN}. The output of the network was a binary mask: a value of 1 is given for pixels predicted in a stone, 0 otherwise. We used the U-Net architecture \cite{20} with a basis of 24 filters of $3 \times 3$ – 24 for the first layer, 48 for the second and so on, as proposed in \cite{21}. The loss function was a combination of binary cross-entropy \cite{22} and Dice loss \cite{23} \cite{24}. The following parameters were employed: input resolution=$256 \times 256$, batch size=1, optimizer=Adam \cite{25}, learning rate=0.001, epoch=200, dropout=0.5 after each block of the descending path. To improve the ability for the network to generalize, the training dataset was expanded through data augmentation (horizontal/vertical flips were applied during training). 

\paragraph{Step $\#$2: Frame-wise image quality control module}

At this point, we had binary masks, which locate relevant stone regions, for the current frame and previously recorded ones (chronological order). From this, a quality control (QC) was subsequently performed, which ensured that:

\begin{enumerate}

\item A stone (or at least a large enough fragment) is present in the current frame. To this end, estimated relevant stone regions (i.e., the region with pixels equal to 1 in the current above-mentioned binary mask) had to cover more than $10 \%$ of the image field-of-view.

\item The current steady frame has not been recorded with flying fragments that can hide the kidney stone or motion of the endoscope, as well as a good still frame from video. To this end, the Dice Similarity Coefficient (DSC) between the stone masks segmented in the current frame and the previous one had to exceed a typical threshold of $0.9$ to ensure a sufficient temporal image stability.

\begin{equation}
\DSC = \frac{2\left|A \cap B \right|}{\left|A \right| + \left|B \right|}
\end{equation}

\noindent where $A$ and $B$ are binary masks estimated in Step $\#1$ in the current and the previous frame, respectively. $A \cap B$ is their intersection and $\left| \cdot \right|$ denotes the cardinality of a set (\emph{i.e.}, the number of voxels).

\end{enumerate}

The processing pipeline proceeded to Step $\#$3 only if both conditions were fulfilled (the current image is discarded otherwise, and the pipeline jumped directly to Step $\#$1 applied on the next frame).

\paragraph{Step $\#$3: Frame-wise identification of stone morphologies}

To predict the stone morphologies on a video frame, a dedicated deep CNN was trained using the above-mentioned collection of annotated high quality images. Irrelevant image regions (see Step $\#$1) were not discarded during training. The deep CNN used as a multi-class classification model (5 classes: Ia, IIb, IIIb, Ia+IIb, Ia+IIIb) was a ResNet-Inception-V2 \cite{26}. The following parameters were employed: input resolution=$256 \times 256$, optimizer algorithm for training=Adam, learning rate=0.001, loss function=categorical cross-entropy, batch size=8, number of epochs=100. To improve the ability for the network to generalize, random combinations of scaling (range=[0,0.3]), rotation (range=[-45,45]°), horizontal/vertical flips, brightness variations (range=[0.2,1]) and shifts (range=[-0.2,0.2] of image width/height) were applied during training.

Once trained, the neural network was used to predict stone morphologies on each frame of the videos individually. Irrelevant image regions (see Step $\#$1) were excluded from the prediction (practically, corresponding pixel intensities were set to 0). 

The processing pipeline proceeds to Step $\#$4 once the last frame of the video completed Step $\#$3 (the pipeline jumped directly to Step $\#$1 applied on the next frame otherwise).

\paragraph{Step $\#$4: ESR from all collected predictions}

At this point, we had a list of predicted morphologies (i.e., one prediction for each frame that fulfilled the QC). Based on these predictions, an ESR decision was done. The following scenarios were investigated:

\begin{enumerate}

\item A majority class emerged (i.e., one of the five classes was present in more than $50 \%$ of the list). The process was stopped and the majority class was marked.

\item No majority class emerged. The presence of a mixed stone is further investigated. The class Ia+IIb (resp. Ia+IIIb) was marked if either one of the 3 classes Ia, IIb (resp. IIIb) or Ia+IIb (resp. Ia+IIIb) was encountered in more than $50 \%$ of the list.

\end{enumerate}

By default, the most represented class in the prediction list was marked.

The marked class was used as the final AESR prediction.

\subsection{Quantitative assessment of the proposed AESR video classifier}

Based on ESR predictions obtained for each video of the validation database, the following test metrics were calculated for each stone type, individually: balanced accuracy, specificity, sensitivity, precision and F1-score \cite{27}. An overall ESR score was then obtained by calculating the mean and the standard deviation of the diagnostic scores over all stone types.

To assess the benefit of each step of the proposed processing pipeline, the statistical analysis was repeated when the trained morphological classifier (i.e., the neural network in Step $\#$3) was applied:

\begin{enumerate}

 \item Without discarding irrelevant regions surrounding the stone within the images, but irrelevant frames excluded by the QC. 

 \item Using original images (no frame excluded and no image regions discarded).

\end{enumerate}

\subsection{Hardware and implementation}

Our test platform was an Intel Xeon E5-2683 2.4 GHz equipped by a GPU Nvidia Tesla P100 with 16 GB of memory. Our implementation was done using Tensorflow 1.4 and Keras 2.2.4.

\section{Results}

\subsection{Characteristics of AI-training cohorts}

The database used to train the network dedicated to stone segmentation (Step $\#$1 of the processing pipeline) included 100 observations (half=stone surface/half=stone sectioned using LASER/10 observations per morphological classes) with the corresponding manually delineated stone. 

The database used to train the network dedicated to morphological identification (Step $\#$3 of the processing pipeline) included 349 annotated observations of stone surface (pure stones: Ia = 191, IIb = 53, IIIb = 29; mixed stones: Ia + IIb = 64, Ia + IIIb = 12) and 236 annotated observations of stone section (pure stones: Ia = 127, IIb = 30, IIIb = 25; mixed stones: Ia + IIb = 31, Ia + IIIb = 23).

\subsection{Diagnostic performance of the proposed ESR video classifier}

The database used to assess the proposed classifier included 71 videos (50 exhibited only one morphological type and 21 displayed two. Pure stones: Ia = 26, IIb = 16, IIIb = 8; mixed stones: Ia + IIb = 13, Ia + IIIb = 8). The video characteristics (duration/number of frames) are summarized in the table \ref{table:video_charac}. Major recorded clinical events are summarized in the table \ref{table:video_contents}. In most of videos ($64 \%$), the stone was not fragmented at the beginning of the recording (figure \ref{fig:Endoscopic_Session}a). In the majority of cases ($47 \%$), a LASER stone fragmentation was performed (figure \ref{fig:Endoscopic_Session}b). $34 \%$ of videos involved stone free prospection sessions (figure \ref{fig:Endoscopic_Session}c). Use of a clamp (figure \ref{fig:Endoscopic_Session}d) and stone fragment removal (figure \ref{fig:Endoscopic_Session}e) were recorded in $6 \%$ and $4 \%$ of videos, respectively. One can note a significant image blurring due to the presence of urine in $4 \%$ of videos, and large blood traces in the stone surface in $1 \%$ of videos.

\begin{table}
\begin{tabular*}{\textwidth}{@{}c*{15}{@{\extracolsep{0pt plus12pt}}c}}
\hline
Video & \multicolumn{5}{c}{Stone types} & Total \\
\cline{2-6}
Characteristics & Ia & IIb & IIIb & Ia+IIb & Ia+IIIb & \\
   \hline
Number of videos $[\#]$ & 26 & 16 & 8 & 13 & 8 & 71 \\
Duration [s] & 150 $\pm$ 122 & 58 $\pm$ 69 & 12 $\pm$ 13 & 92 $\pm$ 92 & 123 $\pm$ 127 & 100  $\pm$ 108 \\
\textit{Mean $\pm$ Std dev [Min-Max]} & [17-445] & [7-266] & [3-43] & [6-309] & [2-324] & [2-445] \\
   \hline
\end{tabular*}
\caption{Characteristics of intra-operative endoscopic videos used for AESR validation.}
\label{table:video_charac}
\end{table}   
   
\begin{table}
\begin{tabular*}{\textwidth}{@{}c*{15}{@{\extracolsep{0pt plus12pt}}c}}
\hline
Interventional & \multicolumn{5}{c}{Stone types} & Total \\
\cline{2-6}
events & Ia & IIb & IIIb & Ia+IIb & Ia+IIIb & \\
\hline
Stone examination before & 16 (62 $\%$) & 13 (81 $\%$) & 2 (25 $\%$) & 11 (85 $\%$) & 3 (43 $\%$) & 45 (64$\%$)\\
 fragmentation $[\# (\%)]$  & & & & \\
LASER stone fragmentation  & 16 (62 $\%$) & 2 (13 $\%$) & 5 (63 $\%$) & 7 (54 $\%$) & 1 (43 $\%$) & 33 (47 $\%$) \\
$[\# (\%)]$  & & & & \\
Stone free prospection & 11 (42 $\%$) & 6 (38 $\%$) & 0 (0 $\%$) & 6 (46 $\%$) & 1 (14 $\%$) & 24 (34 $\%$)\\
$[\# (\%)]$  & & & & \\
Use of a clamp & 3 (12 $\%$) & 0 (0 $\%$) & 0 (0 $\%$) & 3 (23 $\%$) & 1 (14 $\%$) & 7 (10 $\%$)\\
$[\# (\%)]$  & & & & \\
Removal of a stone fragment & 2 (8 $\%$) & 1 (6 $\%$) & 0 (0 $\%$) & 1 (8 $\%$) & 0 (0 $\%$) & 4 (6 $\%$)\\
$[\# (\%)]$  & & & & \\
Presence of urine throughout & 0 (0 $\%$) & 2 (13 $\%$) & 0 (0 $\%$) & 1 (8 $\%$) & 0 (0 $\%$) & 3 (4 $\%$)\\
$[\# (\%)]$  & & & & \\
Presence of blood at the surface of & 0 (0 $\%$) & 1 (6 $\%$) & 0 (0 $\%$) & 0 (0 $\%$) & 0 (0 $\%$) & 1 (1 $\%$)\\
the stone $[\# (\%)]$  & & & & \\
   \hline
\end{tabular*}
\caption{Summary of intra-operative interventional events observed in intra-operative endoscopic videos used for AESR validation. Interventional events are listed in their order of occurrence in the video database (most frequently encountered first).}
\label{table:video_contents}
\end{table}

\begin{figure}[h!]
\begin{minipage}[b]{\linewidth}
\centering
\centerline{\includegraphics[width=\linewidth]{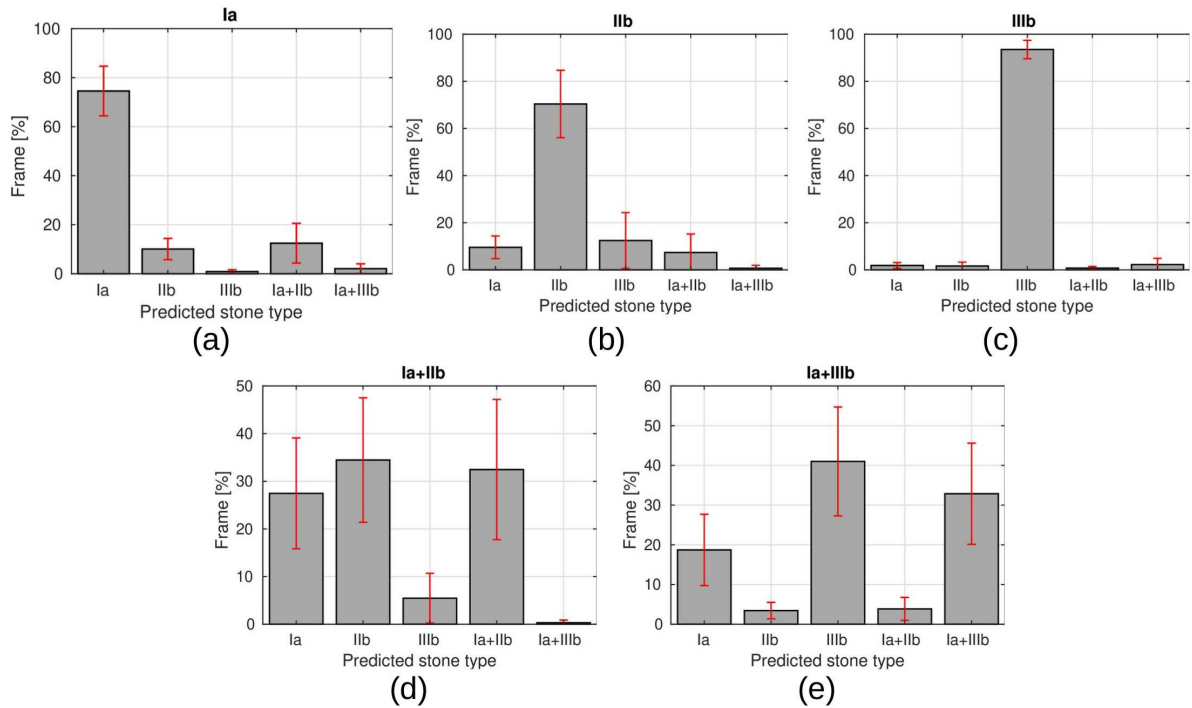}}
\end{minipage}
\caption{Frame-wise analysis of morphologies identified by the implemented classifier. Each panel details predictions collected in videos annotated with a specific stone type (the reference stone type is reported in the panel title). Each bar shows the average percentage of frames in videos for a specific predicted stone type (error bars=standard deviation). In these results, images that don’t fulfil the QC are discarded and regions surrounding the stone are removed during the prediction process.}
\label{fig:Frame_Analysis}
\end{figure}

Using the proposed classifier, the stone type was correctly predicted in average in $~75 \%$ of frames in Ia-annotated videos (figure \ref{fig:Frame_Analysis}a); $~70 \%$ of frames in IIb-annotated videos (figure \ref{fig:Frame_Analysis}b); $~95 \%$ of frames in IIIb-annotated videos (figure \ref{fig:Frame_Analysis}c). One of the 3 classes Ia, IIb or Ia+IIb was predicted in average in $~90 \%$ of frames in (Ia + IIb)-annotated videos (figure \ref{fig:Frame_Analysis}d). The same observation applied for (Ia + IIIb)-annotated videos. Typical videos with superimposed frame-wise AESR results are given in the Supplemental Data.

Over the 71 videos, the percentage of frames that fulfil the QC was $[59 \pm 29] \%$ (pure stones: Ia = $[45 \pm 28] \%$, IIb = $[65 \pm 30] \%$; IIIb = $[87 \pm 13] \%$; mixed stones: Ia + IIb = $[52 \pm 24] \%$, Ia + IIIb = $[73 \pm 20] \%$).
Table \ref{table:scores} details the diagnostic performance of the proposed video classifier for each tested stone type. It can be observed that AESR scores improved when image regions surrounding the stone are discarded from the prediction process (balanced accuracy = $[82 \pm 12] \%$ with, $[88 \pm 6] \%$ without). Along the same line, AESR scores dramatically dropped down when the proposed QC was not used (balanced accuracy = $[54 \pm 6] \%$).

\section{Discussion}

In this study, an automatic computer-aided classifier is proposed to predict \textit{in-situ} the morphology of pure and mixed stones based on intra-operative endoscopic digital videos acquired in a clinical setting. Our approach takes a direct advantage of recent developments in AI-networks dedicated to the region localization in images and to classification tasks.
It must be noticed that the ESR skill requires a learning curve which can be steep, especially when mixed stones morphologies are involved \cite{2} \cite{9} \cite{15}. This limits the translation of ESR to a practical clinical use. The potential advantages of an automated computer-aided tool lie in the generation of reproducible morphological identifications and in the minimization of operator dependency. Ideally, the tool must deliver identical results, either with static images collected by a trained endo-urologist or by someone unfamiliar with stone morphological types. However, the goal of a reliable, automatic and reproducible ESR is, in practice, seriously hindered by the need for a non-intuitive intra-operative manual selection of a steady frame with a sufficient image quality. As reported in table \ref{table:video_contents}, images acquired dynamically on-the-fly during the common practice of an intra-operative endoscopic imaging session are prone to several interventional events such as: instability of the endoscopic video image, stone-free prospection sessions, particles (stone dusting/fragments) flying around in the saline, among others. It must also be underlined that such events are not necessarily present in images used for the AI-training. The occurrence of such events during an intra-operative AESR session may disturb, in turn, the reliability of an AI-model. Our videos-based AESR approach ― which embeds a suitable QC module from frame selection (Step $\#$2 in the proposed processing pipeline) ― provides morphological predictions without resorting to any human intervention for the selection of a steady frame (see table \ref{table:scores}).

\begin{table}
\begin{tabular*}{\textwidth}{@{}c*{15}{@{\extracolsep{0pt plus12pt}}c}}
\hline
Diagnostic performance & \multicolumn{5}{c}{Stone types} & Overall ESR scores \\
\cline{2-6}
indicator & Ia & IIb & IIIb & Ia+IIb & Ia+IIIb & \textit{Mean $\pm$ Std dev}\\
   \hline
\multicolumn{7}{c}{Using the complete pipeline \textit{(QC + irrelevant image regions excluded)}} \\
   \hline
Balanced accuracy [$\%$] & 90 & 86 & 96 & 81 & 85 & 88 $\pm$ 6\\
Sensitivity [$\%$] & 85 & 75 & 100 & 69 & 71 & 80 $\pm$ 13\\
Specificity [$\%$] & 95 & 96 & 92 & 93 & 98 & 95 $\pm$ 2\\
Precision [$\%$] & 92 & 86 & 62 & 69 & 83 & 78 $\pm$ 12\\
F1-score [$\%$] & 88 & 80 & 76 & 69 & 77 & 78 $\pm$ 7\\
   \hline
\multicolumn{7}{c}{No image  regions discarded \textit{(QC applied)}}\\
\hline
Balanced accuracy [$\%$] & 89 & 81 & 97 & 67 & 76 & 82 $\pm$ 12\\
Sensitivity [$\%$] & 96 & 63 & 100 & 38 & 57 & 71 $\pm$ 26\\
Specificity [$\%$] & 82 & 100 & 94 & 95 & 95 & 93 $\pm$ 7\\
Precision [$\%$] & 76 & 100 & 67 & 63 & 57 & 73 $\pm$ 17\\
F1-score [$\%$] & 85 & 77 & 80 & 48 & 57 & 69 $\pm$ 16\\
   \hline
\multicolumn{7}{c}{Using all frames from videos / no image regions discarded}\\
\hline
Balanced accuracy [$\%$] & 50 & 50 & 50 & 58 & 64 & 54 $\pm$ 6\\
Sensitivity [$\%$] & 0 & 0 & 0 & 38 & 100 & 28 $\pm$ 44\\
Specificity [$\%$] & 100 & 100 & 100 & 77 & 29 & 81 $\pm$ 31\\
Precision [$\%$] & 0 & 0 & 0 & 28 & 13 & 8 $\pm$ 12\\
F1-score [$\%$] & 0 & 0 & 0 & 32 & 24 & 11 $\pm$ 16\\
   \hline
\end{tabular*}
\caption{Diagnostic performance of the proposed video classifier. Balanced accuracies, sensitivities, specificities, precisions and F1-scores are shown in percentages for each tested stone type. AESR scores, averaged over the five analysed morphological classes, are emphasized with bold characters (top right column). Scores are also reported i) without discarding irrelevant regions surrounding the stone in the images (irrelevant frames excluded), ii) Using all frames from videos, no image regions discarded.}
\label{table:scores}
\end{table}

While the QC module is able to detect irrelevant image information at the frame level, the proposed pipeline also deals with irrelevant image regions at the pixel level: AESR scores are improved when regions surrounding the stone are discarded within frames, as shown in table \ref{table:scores}. By discarding irrelevant image information at frame and pixel levels, a large majority of frames ($>75 \%$ of the video in average) provided valid predictions in pure stones, as shown in panels (a-c) in figure \ref{fig:Frame_Analysis}. The correct majority class emerged with a high balanced accuracy ($> 86 \%$) for pure stones (table \ref{table:scores}). Interestingly, for Ia stones, wrong predictions were a mixed type with Ia morphology in one of the two morphologies. For IIb stones, IIIb were predicted in 3 videos: one can note blurring induced by urine in one video throughout, large traces of blood in the stone surface in another, and high specular reflections of the light of the endoscope in the third.

It must be underlined that valuable information about stone morphology can be found in the surface, section and nucleus of stones, as reported in several studies [10,11]. As a consequence, mixed morphologies may be revealed during the time-course of the stone fragmentation process.  It can be observed that, in a frame-by-frame basis, one of the two morphologies constituting mixed stones was generally detected by default by our classifier, as shown in panels (d) and (e) in figure 3. The correct stone type was selected in the Step $\#$4 of the processing pipeline with a high balanced accuracy ($> 81 \%$) for mixed stones (table \ref{table:scores}). Interestingly, final ESR predictions always contained one of the two morphologies constituting mixed stones.

Ultimately, it must be underlined that the reliability of the annotated datasets used for AI-training is crucial, since any subjectivity in ESR of the urologist may be unfortunately directly transferred into an AI-model. We believe that the annotation task must rely on the concordance between endoscopic and microscopic examinations, as described in \cite{9} \cite{3} \cite{10} \cite{11}.

\section{Conclusion}

The current study demonstrates that AI applied on digital endoscopic video sequences is a promising tool for collecting morphological information during the time-course of the stone fragmentation process. In particular, this is achievable without resorting to any human intervention for stone delineation or selection of good quality steady frames. To this end, irrelevant image information must be removed from the prediction process at both frame and pixel levels, which is now feasible thanks to the use of AI-dedicated networks. Such a tool may be beneficial to assist endo-urologist visual interpretation of stone morphologies before and during the interventional procedure, which is essential for an aetiological diagnosis of stone disease. 

For a wide adoption of ESR into clinical routine, major efforts must focus on the creation of larger set of data corresponding to specific aetiologies or lithogenic mechanisms, and annotated according to the criteria published [9]. Such an AI-training cohort, with morphological annotations confirmed by both endoscopy and microscopy, is essential to improve AESR scores and to extend the method to a wider range of pure and mixed stones types. 

\section*{Acknowledgment}

Experiments presented in this paper were carried out using the PlaFRIM experimental testbed, supported by Inria, CNRS (LABRI and IMB), Universit\'e de Bordeaux, Bordeaux INP and Conseil R\'egional d'Aquitaine (see https://www.plafrim.fr/). Computer time for this study was provided by the computing facilities MCIA (M\'esocentre de Calcul Intensif Aquitain) of the Universit\'e de Bordeaux and of the Universit\'e de Pau et des Pays de l'Adour.

\section*{References}

\bibliographystyle{dcu}
\bibliography{2022_Estrade_deep_Video_recognition}

\end{document}